\documentclass[10pt,twocolumn,letterpaper]{article}

\usepackage{wacv}              %

\usepackage{multirow}   %
\usepackage[protrusion=true,expansion=true,tracking=false,kerning=true]{microtype}  %

\graphicspath{{figures/}}
\usepackage{placeins}

\definecolor{wacvblue}{rgb}{0.21,0.49,0.74}
\usepackage[pagebackref,breaklinks,colorlinks,allcolors=wacvblue]{hyperref}
\hypersetup{pdftitle={SurgQ-VLM: Question-Conditioned Dual-Stream Compression and Answer-Type-Aware Reinforcement for Surgical Video Question Answering}, pdfauthor={Dhruv Makwana, Ananya Mantravadi}}

\def\wacvPaperID{2553} %
\def\confName{WACV}
\def\confYear{2027}

\title{SurgQ-VLM: Question-Conditioned Dual-Stream Compression and
Answer-Type-Aware Reinforcement for Surgical Video Question Answering}

\author{Dhruv Makwana\\
Centific Global Solutions\\
{\tt\small dhruv.makwana@centific.com}
\and
Ananya Mantravadi\\
Centific Global Solutions\\
{\tt\small ananya.mantravadi@centific.com}
}

\begin{document}
\maketitle
\makeatletter
\AddToShipoutPicture{\ifnum\value{page}=1\relax\AtPageLowerLeft{\put(\strip@pt\dimexpr0.5\paperwidth\relax,\strip@pt\dimexpr0.32in\relax){\makebox[0pt]{\parbox[b]{\textwidth}{\centering\scriptsize
Accepted to the 2027 IEEE/CVF Winter Conference on Applications of Computer Vision (WACV). \copyright\ 2027 IEEE. Personal use of this material is permitted. Permission from IEEE must be obtained for all other uses, in any current or future media, including reprinting/republishing this material for advertising or promotional purposes, creating new collective works, for resale or redistribution to servers or lists, or reuse of any copyrighted component of this work in other works.}}}}\fi}
\makeatother
\setlength{\textfloatsep}{8pt plus 2pt minus 2pt}
\setlength{\floatsep}{6pt plus 2pt minus 2pt}
\setlength{\intextsep}{6pt plus 2pt minus 2pt}
\setlength{\dbltextfloatsep}{8pt plus 2pt minus 2pt}
\makeatletter
\renewcommand\paragraph{\@startsection{paragraph}{4}{\z@}%
  {1.5ex \@plus0.5ex \@minus.2ex}{-1em}%
  {\normalfont\normalsize\bfseries}}
\makeatother

\begin{abstract}
Reasoning about surgical video --- which phase is underway, which instruments act, and when events occur --- underpins operative documentation, workflow analysis, and skill assessment. Yet surgical visual question answering is still largely built on single frames and closed labels that cannot express a duration, an ordering, or a count, and the video models that lift the frame limit typically compress every clip the same way regardless of the question, discarding the temporal evidence the answer needs. Worse, the answers are structurally heterogeneous, spanning five incompatible structures --- binary, classification, multi-label, counting, and sentence-temporal --- that no single training signal scores uniformly well. We propose SurgQ-VLM, a two-stage model for surgical video question answering. Stage~1 trains a question-conditioned dual-stream (slow/fast) compressor for a LoRA-adapted Qwen2.5-VL-7B: cross-attention conditions both streams on the question, and each clip's visual tokens shrink $49{\times}$ from the raw ViT patches ($11{\times}$ vs.\ post-merge backbone tokens). Stage~2 post-trains with GRPO under an answer-type-aware reward that scores each question on the metric matching its answer structure (micro-F1 for label sets, normalised error for counts, a content--timing blend for temporal windows). On a sealed 21{,}840-question test split over Cholec80 and CholecT50, SurgQ-VLM reaches 0.779 keyword accuracy and a $10{\times}$ BLEU-4 gain over zero-shot; GRPO adds 2.5\,pp K-Acc (2.0\,pp attributable to the answer-type-aware reward), concentrated in the structurally hard types, with answer format unchanged (ROUGE-L). Code is available at \url{https://github.com/DhruvMakwana/SurgQ-VLM}.

\end{abstract}

\section{Introduction}
\label{sec:intro}

\begin{figure}[t]
  \centering
  \includegraphics[width=\linewidth,viewport=72 320 392 707,clip]{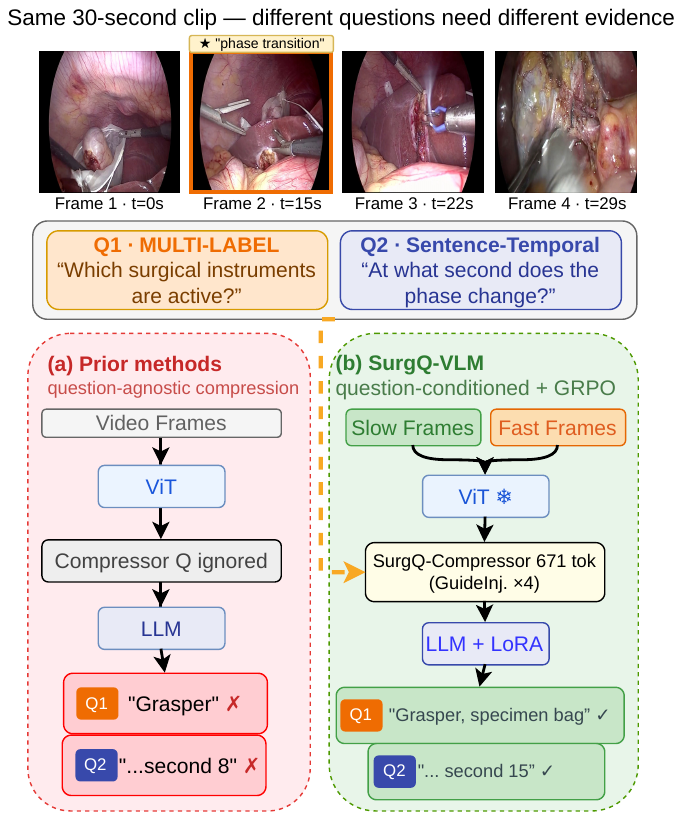}
  \caption{The same 30-second laparoscopic clip requires different temporal evidence for different questions. Prior methods~(\emph{a}) compress without question context, producing incomplete or mistimed answers. SurgQ-VLM~(\emph{b}) conditions both streams on the question via four independent GuideInjectors, then refines with GRPO's answer-type-aware reward.}
  \label{fig:teaser}
  \vspace{-4mm}
\end{figure}

Automated reasoning over surgical video underpins a growing set of clinically motivated tasks --- operative documentation, operating-room workflow analysis and scheduling, and objective skill assessment --- each of which depends on knowing not only \emph{what} happens in a procedure but \emph{in what order} and \emph{when}~\cite{surgicaldatascience, cholec80}. This evidence is inherently temporal: a phase begins at one second and ends at another, an instrument first appears just before a transition, a clip contains some number of distinct actions. Yet surgical visual question answering (VQA) is still built on single frames and closed label sets~\cite{surgicalvqa, pitvqa, ssgvqa}, which cannot express a duration, an ordering, or a count by construction.

Video vision-language models (VLMs)~\cite{videollava, videochatgpt, llavaonevision} remove the single-frame limit, but they compress every clip the same way regardless of the question, so the temporal evidence a question needs is often the first evidence the compressor discards (\cref{fig:teaser}\emph{a}). Surgery makes this acute: from the same pixels, the same clip may be asked for a phase label, an instrument set, a count, or the second at which a transition occurs --- answers spanning five incompatible structures (binary, classification, multi-label, counting, sentence-temporal) that no single training signal scores correctly.

Prior work addresses only fragments of this. Surgical video VLMs that move beyond single frames still train with supervised fine-tuning alone~\cite{surgvidlm, llavasurg}; question-conditioned compressors condition a single unified stream~\cite{hicom}; dual-rate samplers keep a slow and a fast stream but compress both question-agnostically~\cite{slowfastllava, slowfast}; and reinforcement post-training for medical video normalises one reward \emph{across datasets}~\cite{medgrpo} rather than across the answer structures inside a single task. Three gaps remain. \textbf{(i)} Question-agnostic compression discards the temporal evidence a question needs before the language model ever sees it. \textbf{(ii)} SFT plateaus on the structurally hard question types (multi-label, sentence-temporal). \textbf{(iii)} A single scalar RL reward conflates the five incompatible answer structures. We propose \textbf{SurgQ-VLM} (\cref{fig:teaser}\emph{b}): Qwen2.5-VL~\cite{qwen25vl} with LoRA~\cite{lora} and a \emph{per-stream} question-conditioned dual-stream compressor that extends HICom~\cite{hicom} and SlowFast-LLaVA~\cite{slowfastllava}, post-trained through a two-stage supervised fine-tuning (SFT) $\to$ GRPO~\cite{grpo} pipeline whose continuous reward dispatches on answer structure. Surgery is our testbed; all experiments are on laparoscopic cholecystectomy. The compressor design and answer-type reward family make no procedure-specific assumptions --- ordered phases, acting instruments, and time-stamped events recur across procedural domains --- but cross-procedure generalization is untested and a concrete next step.

\noindent\paragraph{Contributions: }
\begin{itemize}
  \item \textbf{An answer-type-aware continuous reward family for GRPO.} Five distinct $R_1$ functions dispatch on answer structure across eighteen Q-types, weighted by Q-type inverse frequency. GRPO lifts K-Acc $+2.5$\,pp over SFT; gains concentrate on the three structurally hard types (multi-label, counting, sentence-temporal) while binary and classification remain flat, with improvement in content rather than surface form.
  \item \textbf{A question-conditioned dual-stream compressor.} Four independent GuideInjectors shape a 30-second clip into four separately addressable token groups (slow/fast $\times$ local/global) within a fixed 671-token budget --- a 49$\times$ compression of the clip's 32{,}805 ViT patch tokens. Question conditioning is the load-bearing element (removing it costs 10.6\,pp K-Acc, 21.2\,pp sentence-temporal); dual-streaming adds a further $+0.9$\,pp Primary and enables the per-stream analysis of \cref{sec:exp-stream}.
  \item \textbf{A multi-task surgical video-QA benchmark} of 80{,}636 QA pairs automatically generated from the per-frame phase, instrument, and triplet annotations of Cholec80 and CholecT50~\cite{cholec80, cholect50}, spanning eighteen question types and five answer structures across 6{,}701 clips, with a sealed 1{,}914-clip / 21{,}840-QA test split.
\end{itemize}

On the sealed test split, SurgQ-VLM reaches $0.779$ K-Acc and a $10\times$ BLEU-4 gain over the Qwen2.5-VL-7B zero-shot baseline; the MedGRPO model~\cite{medgrpo} (uAI-NEXUS MedVLM) scores $0.398$ zero-shot out-of-domain. The GRPO gain concentrates in the open-ended answer types (multi-label, counting, sentence-temporal), while surface form stays fixed (ROUGE-L $p{=}0.98$). Code and trained weights are public.

\section{Related work}
\label{sec:related}

\paragraph{Video VLMs and temporal compression.}
Video VLMs share a common structure: a frozen vision encoder produces per-frame tokens, a compressor reduces them to a budget the language model can consume, and the language model is fine-tuned on video--text pairs~\cite{videollava,videochatgpt,llavaonevision}. Token compression is the design axis where these models differ. ChatUni-VI~\cite{chatunivi} clusters dynamic visual tokens, MovieChat~\cite{moviechat} maintains sparse memory banks, and LongVU~\cite{longvu} prunes by cross-frame similarity --- but all remain question-agnostic: the same tokens are produced for every query on a clip. Two recent threads push against this. SlowFast-LLaVA-1.5~\cite{slowfastllava} borrows the SlowFast split~\cite{slowfast} to trade spatial against temporal resolution across two streams, and SlowFocus~\cite{slowfocus} localises the question-relevant temporal segment and samples it at higher density for fine-grained temporal reasoning. HICom~\cite{hicom} instead conditions a single-stream compressor on the question, pooling patches relevant to the query. QG-VTC~\cite{qgvtc} does the same for static images. SurgQ-VLM differs from both: it conditions each stream independently through four GuideInjectors, producing four question-shaped token groups rather than a single conditioned representation.

\vspace{-2pt}

\paragraph{Surgical and medical VLMs.}
Surgical-VQA~\cite{surgicalvqa} introduced transformer-based surgical question answering, releasing the Cholec80-VQA and EndoVis-18-VQA benchmarks by bootstrapping QA pairs from per-frame phase and instrument annotations. A line of single-frame follow-up extends this template: GPT-2 decoding~\cite{surgicalgpt}, gated fusion~\cite{catvil}, localised answering~\cite{surgicalvqla,surgicalvqlapp}, pituitary adaptation~\cite{pitvqa}, and scene-graph priors~\cite{ssgvqa}. SurgViVQA~\cite{surgvivqa} moves to short clips with a GPT-2 decoder and introduces the keyword-accuracy protocol we adopt. Recent surgical LVLMs --- LLaVA-Surg~\cite{llavasurg}, Surgical-LLaVA~\cite{surgicalllava}, and Surgical-LVLM~\cite{surgicallvlm} adapt general-purpose VLM assistants to the surgical domain, and general medical foundation models such as LLaVA-Med~\cite{llavamed}, VILA-M3~\cite{vilam3}, Med-PaLM M~\cite{medpalmm}, and Med-Gemini~\cite{medgemini} target broader biomedical tasks without procedural temporal structure. The closest work moves to video: SurgVidLM~\cite{surgvidlm} targets multi-grained surgical video understanding, and Surgery-R1~\cite{surgeryr1} adds reinforcement fine-tuning to surgical visual question-localised answering, localising answers on robotic-surgery frames rather than scoring open-ended temporal answers.

\vspace{-3pt}

\paragraph{Reinforcement learning for VLM post-training.}
Reinforcement learning has become a standard post-training step for reasoning in language models since GRPO~\cite{grpo}, which replaces a learned value function with group-relative returns and enables on-policy updates without a critic. VLMs followed quickly. For video, Time-R1~\cite{timer1} formulates temporal grounding as a GRPO problem with an IoU-based reward on predicted time spans, and Video-R1~\cite{videor1} introduces T-GRPO, a temporal variant of GRPO that contrasts rollouts on ordered versus shuffled frames to teach reasoning-over-time. In the medical domain, MedVLM-R1~\cite{medvlmr1} and Med-R1~\cite{medr1} post-train image VLMs with GRPO for medical image reasoning, each applying a single reward function per sample regardless of answer structure. Closest to our setting, MedGRPO~\cite{medgrpo} applies multi-task GRPO to heterogeneous medical \emph{video} data, normalising rewards across eight datasets so that imbalanced reward scales do not collapse training. These methods address heterogeneity \emph{across datasets or tasks} with one reward function per sample; none dispatch the reward on the \emph{answer structure within a task}.
\section{Method}
\label{sec:method}

\begin{figure*}[!t]
  \centering
  \includegraphics[width=\linewidth]{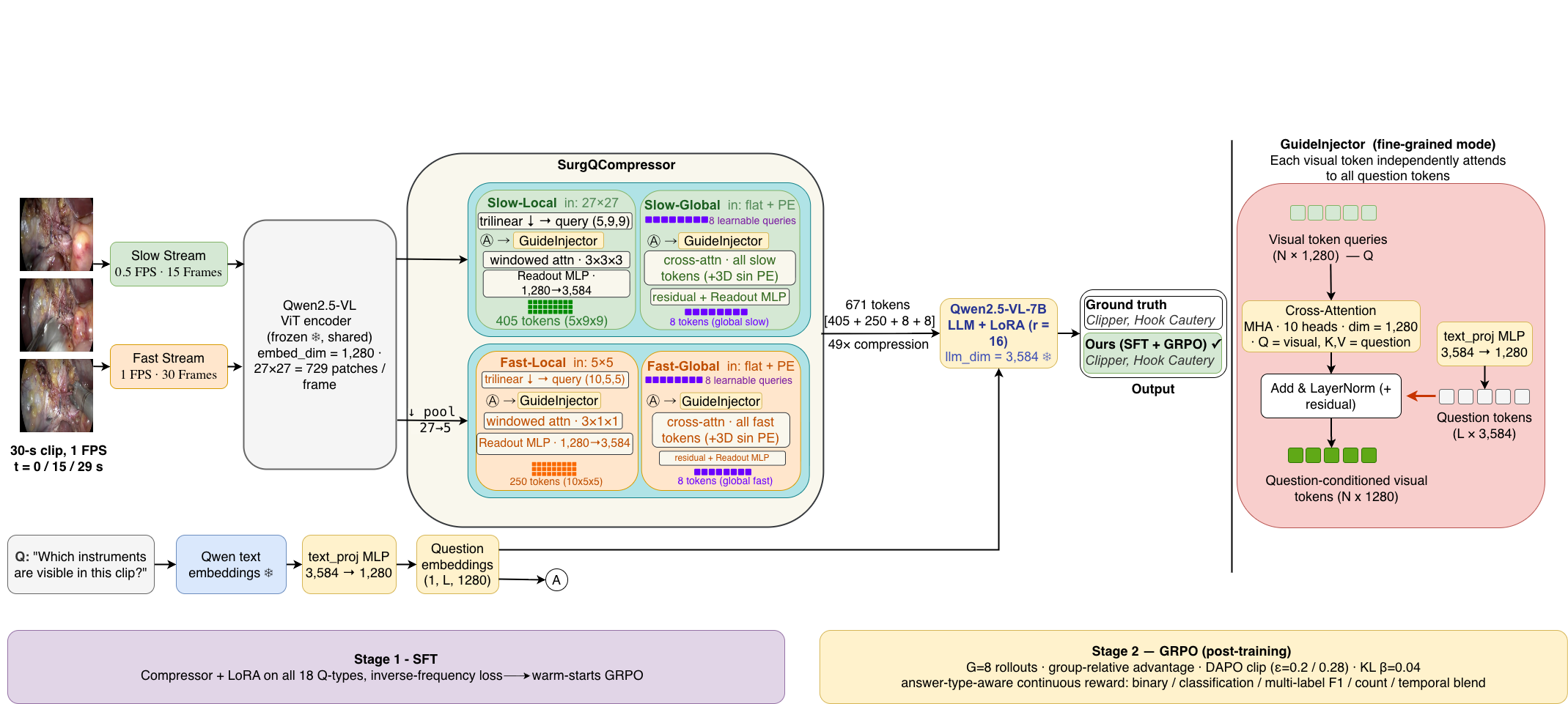}
  \caption{%
    \textbf{SurgQ-VLM pipeline.}
    Dual-stream sampling (slow: 0.5\,FPS/15f; fast: 1.0\,FPS/30f) is encoded by a frozen Qwen2.5-VL ViT, then compressed by four question-conditioned GuideInjectors into 671 tokens (405 slow-local $+$ 250 fast-local $+$ 16 global) before a LoRA-adapted LLM.
    Blue: frozen; green/orange: slow/fast streams; purple: global queries; gold: GuideInjectors. Bottom panel: two-stage training.
  }
  \label{fig:architecture}
\end{figure*}

\Cref{fig:architecture} shows the SurgQ-VLM architecture. A 30-second laparoscopic clip is sampled at two rates, encoded by a frozen vision encoder, and compressed by a question-conditioned dual-stream compressor into a fixed 671-token bundle before being decoded by a LoRA-adapted language model. Prior question-conditioned video compressors apply conditioning to a single unified stream~\cite{hicom}; dual-rate samplers process two streams without per-stream conditioning~\cite{slowfastllava,slowfast}. Our contribution is \emph{per-stream question conditioning}: a GuideInjector --- a cross-attention block, adapted from HICom~\cite{hicom}, in which each visual token queries the question's token embeddings (as keys and values) to absorb question-specific signal --- is applied independently to both the slow and fast streams, so each retains the temporal character of its sampling rate while attending to the question.

Training proceeds in two stages. Stage~1 (SFT) equips the model with surgical domain vocabulary and structured answer format through maximum-likelihood supervision; without it, the policy explores an unstructured output space and open-ended question types degrade. Stage~2 (GRPO) refines the SFT checkpoint using a continuous reward that provides gradient signal where cross-entropy cannot. We describe each component in turn.

\subsection{Dual-stream visual encoding}
\label{sec:method-setup}

A clip is a 30-second segment of laparoscopic video paired with a question $q$ and target answer $a$. Eighteen question types span five answer structures: binary, classification, multi-label, counting, and sentence-temporal. Clips are sampled at two rates chosen to match the temporal scales of the two evidence types the task requires. The \emph{slow stream} takes one frame every two seconds (0.5\,FPS, 15 frames), preserving the spatial detail needed to identify fine anatomical structures --- the hepatocystic triangle, gallbladder attachment site, and clip placement --- that change slowly within a phase and are critical for phase classification and instrument identification. The \emph{fast stream} takes one frame per second (1.0\,FPS, 30 frames), capturing instrument state transitions, action-boundary events, and triplet changes that occur on second-level timescales. Both streams pass through the frozen Qwen2.5-VL-7B~\cite{qwen25vl} vision transformer (ViT), producing $27\!\times\!27$ pre-merger patch tokens per frame. A 30-second clip therefore yields $45\!\times\!729 = 32{,}805$ patch tokens (15 slow $+$ 30 fast frame instances; the slow stream re-samples the clip's 30 distinct frames), which are routed into the compressor.

\vspace{-2pt}

\subsection{Local compression}
\label{sec:method-local}

The local compressor processes each stream at its own spatial and temporal resolution. For the slow stream, we pool $15\!\times\!27\!\times\!27$ patch tokens to $5\!\times\!9\!\times\!9$ (a $3 \!\times$ spatial reduction): the $9\!\times\!9$ readout discriminates adjacent anatomical structures while keeping the per-stream token count tractable. For the fast stream, we pool $30\!\times\!27\!\times\!27$ to $10\!\times\!5\!\times\!5$ (a $5.4 \!\times$ reduction): at faster temporal density the discriminating signal is instrument presence and coarse motion, so we trade spatial fidelity for temporal coverage. Each readout slot attends over its own spatio-temporal window at the original $27\!\times\!27$ resolution through a windowed attention block (kernels $3\!\times\!3\!\times\!3$ for slow, $3\!\times\!1\!\times\!1$ for fast), so pooling summarises rather than discards. Before readout, each pooled visual token slot is conditioned on the question. We embed $q$ with Qwen2.5-VL's frozen token-embedding layer to produce a sequence of question token embeddings; these are projected from LLM dimension (3,584) to ViT dimension (1,280) via a learned linear layer before serving as keys and values in the cross-attention. The full grid of visual token queries then passes through a GuideInjector: each visual token independently cross-attends over all question token embeddings (fine-grained per-token conditioning, 10-head MHA) --- the visual token as query, projected question tokens as keys and values. Unlike HICom~\cite{hicom}, where an instruction-conditioned token queries its own spatial group, every spatial position here reads the entire question, so a region containing a clipper attends to the word \emph{clipper}. (A coarser AdaLN-style injection~\cite{dit} would lose this per-location specificity.) Each compressor block closes with a two-layer readout MLP (Linear\,1{,}280$\!\to\!$3{,}584, GELU, Linear\,3{,}584$\!\to\!$3{,}584) that projects compressed tokens back to LLM dimension. The slow and fast local paths each have their own GuideInjector and readout instances. The local stage yields $5\!\times\!9\!\times\!9 = 405$ slow-local tokens and $10\!\times\!5\!\times\!5 = 250$ fast-local tokens.

\vspace{-2pt}

\subsection{Global queries}
\label{sec:method-global}

Local compression processes each stream inside bounded spatio-temporal windows, which means evidence that spans multiple windows --- a phase transition falling at a window boundary, or an instrument appearing briefly in a single frame --- can be missed. The global path addresses this by attending over the entire flattened clip rather than any window. Eight learnable queries per stream cross-attend over all ViT patch tokens from that stream's frames, building a fixed-size clip-level summary that is complementary to the local path. As with the local paths, the slow and fast global paths stay separate: slow-stream queries capture holistic spatial context (anatomy layout, phase-level appearance) and fast-stream queries the temporal narrative (instrument appearances and action changes), whereas a single shared set over all 45 frames would collapse these into one undifferentiated representation. Queries are zero-initialised and diverge in training, as each occupies a distinct LLM sequence position and so receives a distinct gradient; the concatenated ViT patch tokens receive a 3D sinusoidal position embedding so the queries can resolve temporal and spatial position across the entire clip. Each stream has its own GuideInjector and cross-attention block — four independent GuideInjector instances in total across local and global paths, each with separate weights — completing the per-stream conditioning. The four token groups then concatenate as \texttt{slow\_local} $\Vert$ \texttt{fast\_local} $\Vert$ \texttt{slow\_global} $\Vert$ \texttt{fast\_global}, yielding $405 + 250 + 8 + 8 = 671$ visual tokens per clip --- a $49\!\times$ compression of the 32{,}805 raw patch tokens --- with all question-specific information encoded inside the compressor rather than after it.

\vspace{-2pt}

\subsection{Stage~1: Supervised fine-tuning}
\label{sec:method-sft}

The compressor and LoRA~\cite{lora} adapters on the Qwen2.5-VL-7B~\cite{qwen25vl} LLM are trained jointly on the full 18-Q-type mixture. LoRA targets both attention and MLP modules (\texttt{q,k,v,o,gate,up,down\_proj}) at rank 16, $\alpha{=}32$. After PEFT wrapping we re-freeze all \texttt{visual.*} parameters. Freezing the ViT preserves its large-corpus prior while the compressor and LoRA adapters specialise for surgical video. We apply Q-type inverse-frequency loss weighting so that binary questions --- which dominate the raw count --- do not swamp the long-tail classes. Optimiser is 8-bit AdamW at learning rate $2\!\times\!10^{-4}$, batch size 8 with gradient accumulation 8 (effective batch 64), three epochs on the SFT split; validation loss plateaus by epoch two, confirming the policy reaches SFT capacity before Stage~2 begins. Unlike prior conditional compressors that pre-train injection on a separate instruction corpus~\cite{hicom}, the question-conditioning here learns entirely within Stage~1.

\subsection{Stage~2: GRPO post-training}
\label{sec:method-grpo}

SFT establishes structured answer format and domain vocabulary, but token-level cross-entropy encodes no notion of count distance: nothing in the loss prefers ``2 actions'' over ``0 actions'' when the target is ``3''. GRPO~\cite{grpo} provides metric-aligned, sequence-level feedback that closes this gap. The compressor is frozen at this stage; only the LoRA adapters receive gradients. For each prompt we sample $G{=}8$ rollouts from the policy and compute group-relative advantages. The reference model is the SFT checkpoint, re-materialised by swapping LoRA weights. We apply DAPO-style decoupled clipping~\cite{dapo} with $\varepsilon_{\text{low}}{=}0.2$ and $\varepsilon_{\text{high}}{=}0.28$ (DAPO paper defaults), which allows asymmetric pressure on exploration. We retain a small KL-divergence penalty ($\beta{=}0.04$) to the reference. Dynamic sampling discards prompts with reward std below $10^{-8}$ (all rollouts scored equally) after $2\!\times$ oversampling. Optimiser is 8-bit AdamW at $5\!\times\!10^{-7}$, batch size 4 with gradient accumulation 2 (effective batch 8), \texttt{max\_new\_tokens}{=}384 during rollout generation, one epoch over the GRPO training split.

\subsection{Answer-type-aware continuous reward}
\label{sec:method-reward}

A single scalar tied to exact-match correctness would conflate eighteen question types with incompatible answer structures: multi-label sets need F1, counts need normalised error, temporal windows need overlap. We define
\[
  R \;=\; R_1 \;+\; 0.3\,R_2 \;+\; 0.05\,R_3 \;-\; 0.3\,R_{\text{len}},
\]
each term scored in $[0,1]$ before weighting. $R_1$ (task correctness) dispatches on the answer-structure group, returning a value in $[0,1]$:
\begin{itemize}
  \item \textbf{Binary}: indicator on normalised yes/no match ($0$ or $1$).
  \item \textbf{Classification}: $1$ if at least half the gold answer's content words appear in the prediction, else $0$ --- multi-word labels (e.g., \emph{calot triangle dissection}) count as retrieved when most of their tokens appear.
  \item \textbf{Multi-label} (instrument sets, verb--instrument--target triplets): micro-F1 over the predicted item set --- precision and recall are computed over all predicted and gold labels pooled together, so a model that recovers some labels from a long triplet list earns partial credit rather than zero.
  \item \textbf{Counting}: $1 - \min(1, |\hat c - c|/\max(c,1))$, normalising absolute count error by the ground-truth count $c$.
  \item \textbf{Sentence-temporal}: a content--timing blend $0.2\cdot\tfrac{k+\tau}{2} + 0.8\,(k\cdot\tau)$, where $k$ is continuous keyword accuracy on the answer text and $\tau$ is temporal accuracy within $\pm2$\,s. The dominant multiplicative term grants credit only when both the named content and the timing are right; the small additive floor keeps a gradient when one side is zero.
\end{itemize}
$R_2$ is a soft format check on the reasoning trace, rewarding \texttt{<think>}/\texttt{</think>} delimiters (one-third each) and at least two ``Step $N$'' markers (one-third), while subtracting a penalty for annotation-lookup phrasing (``the annotation'', ``frame index $N$'', \texttt{*\_per\_frame} field names) that early rollouts discovered as a reward shortcut. SFT targets contain no reasoning traces; the trace format emerges during GRPO under $R_2$. $R_3$ applies only to the six sentence-temporal Q-types (Q03, Q04, Q08, Q09, Q16, Q18): it rewards at least one numeric second or frame reference in the reasoning trace ($+0.5$) and temporal vocabulary such as \emph{before/after/transition/begins} ($+0.5$); non-temporal Q-types receive $R_3{=}1$ by default. $R_{\text{len}}$ is a linear overlong penalty that prevents the policy from drifting into long rollouts as the context budget approaches. Each prompt's policy-gradient loss is scaled by the same Q-type inverse-frequency weight used in Stage~1 (applied after group-relative advantage normalisation), so rare Q-types contribute as much gradient as frequent ones despite far fewer prompts. Coefficients were calibrated on a short pilot run (5\% of training data, $G{=}4$): the $R_3$ weight was reduced from $0.2$ to $0.05$ after the pilot revealed the model citing annotation-schema field names to gain temporal reward --- a pattern $R_2$'s annotation-lookup penalty also penalises, making a large $R_3$ redundant. $R_2$ and $R_{\text{len}}$ share the same weight ($0.3$): symmetric auxiliary coefficients beneath the task-correctness weight ($R_1$, $1.0$).

\begin{table}[h]
  \centering
  \footnotesize
  \setlength{\tabcolsep}{2pt}
  \caption{Benchmark distribution across the five answer structures. Q-type inverse-frequency weighting compensates for the imbalance between binary and counting questions.}
  \label{tab:dataset}
  \begin{tabular}{p{1.9cm}p{3.0cm}rc}
    \toprule
    Answer structure & Q-types & Test QA & Metric \\
    \midrule
    Binary           & Q02, Q06, Q07, Q13, Q17      & 7{,}120 & Bin. accuracy \\
    Classification   & Q01, Q10, Q14, Q15           & 3{,}522 & K-Acc \\
    Multi-label      & Q05, Q11                     & 2{,}446 & Micro-F1 \\
    Counting         & Q12                          & 542     & Exact-match \\
    S-temporal       & Q03, Q04, Q08, Q09, Q16, Q18 & 8{,}210 & K-Acc, T-Acc$_{\pm2}$ \\
    \midrule
    \textbf{Total}   & Q01--Q18 (18 types)          & \textbf{21{,}840} & --- \\
    \bottomrule
  \end{tabular}
\end{table}

\section{Experiments}
\label{sec:experiments}

\subsection{Setup}
\label{sec:exp-setup}

\begin{table*}[t]
  \centering
  \footnotesize
  \setlength{\tabcolsep}{3pt}
  \caption{Results on the sealed 21{,}840-QA test split; our final model in bold. \textsuperscript{$\dagger$}Backbone-only row also serves as the architecture ablation. \texttt{n/a}: not logged for baseline/ablation runs; \texttt{---}: not applicable.}
  \label{tab:main}
  \begin{tabular}{lllllccccccc}
    \toprule
    Method & Backbone & Type & \#Fr. & \#Tok. & K-Acc & Primary & BLEU-4 & ROUGE-L & METEOR & CIDEr & T-Acc$_{\pm2}$ \\
    \midrule
    \multicolumn{12}{@{}l}{\emph{Zero-shot}} \\
    Qwen2.5-VL (backbone)            & 7B & ZS   & 45 & 7{,}605  & 0.220 & 0.293 & 0.057 & 0.255 & 0.193 & 0.260 & 0.510 \\
    uAI-NEXUS MedVLM~\cite{medgrpo}  & 7B & ZS   & 8  & 1{,}352  & 0.398 & 0.376 & 0.038 & 0.214 & 0.183 & 0.125 & 0.338 \\
    \midrule
    \multicolumn{12}{@{}l}{\emph{Fine-tuned on our SFT split}} \\
    LLaVA-OneVision-7B~\cite{llavaonevision} & Qwen2    & +LoRA    & 8  & 1{,}568 & 0.653 & 0.621 & 0.481 & 0.759 & 0.703 & n/a & n/a \\
    SurgVidLM~\cite{surgvidlm}               & Qwen2-VL & +LoRA    & 30 & 540     & 0.740 & 0.710 & 0.554 & 0.751 & 0.710 & n/a & n/a \\
    SurgViVQA~\cite{surgvivqa}               & GPT-2    & full FT  & 8  & ---     & 0.132 & 0.132 & 0.097 & 0.297 & 0.218 & n/a & n/a \\
    \midrule
    \multicolumn{12}{@{}l}{\emph{Ablation + ours}} \\
    Qwen2.5-VL (no compressor)\textsuperscript{$\dagger$} & 7B & +LoRA & 45 & 7{,}605 & 0.746 & 0.711 & 0.564 & 0.707 & 0.732 & 5.158 & 0.842 \\
    Single-stream (slow-only) SFT      & 7B & ours & 15 & 671     & 0.750 & 0.713 & 0.555 & 0.812 & n/a   & n/a   & 0.875 \\
    SurgQ-VLM (SFT)                    & 3B & ours & 45 & 671 & 0.736 & 0.709 & 0.561 & 0.797 & 0.722 & 5.211 & 0.839 \\
    SurgQ-VLM (SFT)                    & 7B & ours & 45 & 671 & \underline{0.754} & \underline{0.722} & \underline{0.576} & \underline{0.803} & \underline{0.740} & \underline{5.328} & \underline{0.874} \\
    \textbf{SurgQ-VLM\,+\,GRPO (ours)} & 7B & ours & \textbf{45} & \textbf{671} & \textbf{0.779} & \textbf{0.743} & \textbf{0.586} & \textbf{0.803} & \textbf{0.760} & \textbf{5.442} & \textbf{0.888} \\
    \bottomrule
  \end{tabular}
\end{table*}

Our benchmark totals 80{,}636 QA pairs automatically generated from the per-frame phase, instrument, and triplet annotations of Cholec80 and CholecT50~\cite{cholec80, cholect50}, spanning 6{,}701 thirty-second clips of laparoscopic cholecystectomy. To the best of our knowledge, it is the first surgical video-QA benchmark to jointly span five answer structures within a single procedure; prior work addresses at most closed-label classification~\cite{surgicalvqa, pitvqa}. We partition at the video level, canonicalising the 45 surgeries shared between the two datasets so no surgery crosses splits (supp.\ \S A.1): train (52{,}100 QA / 4{,}212 clips), validation (6{,}696 QA / 575 clips), and a sealed test split of 21{,}840 QA pairs over 1{,}914 clips (held out from all training and hyper-parameter decisions; used only for inference-time evaluation of frozen models). QA pairs come from deterministic rule-based templates over per-frame annotations --- no LM involved --- with five surface-form phrasings per Q-type (Tab.~A1, supplementary). The benchmark and weights are released under CC BY-NC-SA 4.0, inherited from both source datasets, and the code under MIT.

\Cref{tab:dataset} shows the distribution across answer structures; the benchmark is naturally imbalanced, motivating the inverse-frequency weighting in both training stages. The test split spans all eighteen Q-types described in \cref{sec:method-setup}. Decoding is greedy with \texttt{max\_new\_tokens=192} (GRPO rollouts use 384; at evaluation 192 suffices for all answer structures). All fine-tuned baselines train on the same SFT split with their released recipes and, where applicable, our LoRA hyper-parameters.

\emph{Metrics.} \emph{K-Acc} is keyword-presence accuracy --- the fraction of gold keywords present in the predicted answer~\cite{surgvivqa}; precision-soft by design (extra output is not penalised, preserving reasoning traces), it is reported alongside ROUGE-L and BLEU-4 to cover length sensitivity. Our \emph{Primary} metric is the $N$-weighted average of binary accuracy, classification K-Acc, multi-label micro-F1, count exact-match, and sentence-temporal K-Acc (its $\pm2$\,s temporal accuracy saturates and is reported separately: \cref{tab:main}, supp.\ \S A.7), scoring each answer structure on its own terms; because binary questions dominate (32.6\% of test vs.\ counting at 2.5\%), \cref{tab:perqtype} disaggregates by group. For cross-paper comparison we report BLEU-4, ROUGE-L, METEOR, CIDEr, T-Acc$_{\pm2}$, and per-clip inference latency ($1.7{\pm}0.6$\,s/clip excluding video decoding, 21.2\,GB peak on A100). Headline cells carry 95\% bootstrap CIs (10{,}000 resamples; half-widths $\leq 0.5$\,pp on accuracy columns); paired significance tests use permutation with the same count; reported figures are from a single training run per configuration, except the headline GRPO result, verified across three seeds (\cref{sec:exp-ablation}).

\subsection{Main results}
\label{sec:exp-main}

\Cref{tab:main} tells a story of three compounding improvements. The zero-shot backbone produces 0.220 K-Acc --- fluent surgical English, hallucinated content, no structured format. Surgical SFT with the dual-stream compressor is the single biggest jump: K-Acc rises to 0.754 and BLEU-4 from 0.057 to 0.576, a $10\times$ gain, as the model acquires domain vocabulary and the five answer structures. The backbone-only ablation confirms the compressor's contribution: the same SFT recipe without it trails on every metric (BLEU-4 0.564 vs.\ 0.576, ROUGE-L 0.707 vs.\ 0.803, T-Acc 0.842 vs.\ 0.874). GRPO then adds the final $+2.5$\,pp to 0.779 K-Acc, with gains concentrated on the structurally hard answer types (\cref{tab:perqtype}). 

\begin{table}[h]
  \centering
  \footnotesize
  \setlength{\tabcolsep}{5pt}
  \caption{GRPO gains by answer structure (SFT$\,{\to}\,$SFT+GRPO, sealed test). \textbf{Bold}: improvement $>$1\,pp; $\dagger$: Q13 regression ($-3.9$\,pp) discussed in text. Counting has $n{=}542$; its $\Delta$ is within sampling error. Sentence-temporal scored by K-Acc; its saturating T-Acc$_{\pm2\text{s}}$: \cref{tab:main}, supp.\ \S A.7.}
  \label{tab:perqtype}
  \begin{tabular}{l l r r r}
    \toprule
    Answer structure & Metric & SFT & GRPO & $\Delta$ \\
    \midrule
    Binary$^\dagger$ & B-Acc & 0.727 & 0.727 & $0.000$ \\
    Classification   & K-Acc & 0.818 & 0.818 & $0.000$ \\
    Multi-label      & F1    & 0.766 & \textbf{0.778} & $+$\textbf{0.012} \\
    Counting         & Exact & 0.328 & \textbf{0.354} & $+$\textbf{0.026} \\
    Sent.-temporal   & K-Acc & 0.690 & \textbf{0.741} & $+$\textbf{0.051} \\
    \midrule
    \emph{Primary (wtd.\ avg.)} & --- & \emph{0.722} & \emph{0.743} & $+$\emph{0.021} \\
    \bottomrule
  \end{tabular}
\end{table}

\begin{table*}[t]
  \centering
  \footnotesize
  \setlength{\tabcolsep}{4pt}
  \caption{Core ablations on the 7B SurgQ-VLM, grouped by the component removed. Every removal lowers the headline K-Acc below the full model (0.779); the $-$Compressor and $-$GRPO rows are SFT-only (SFT reference: 0.754). Group columns report per-group K-Acc (structure metrics: \cref{tab:perqtype}). $\star$ central contribution. The ``$-$Question conditioning'' row is a dual-stream unconditioned baseline (SlowFast-LLaVA-style~\cite{slowfastllava}) at matched 671-token budget.}
  \label{tab:ablation}
  \begin{tabular}{llccccccc}
    \toprule
    Variant & Removed & K-Acc & Binary & Classif. & Multi-lbl & Count & S-Temp & ROUGE-L \\
    \midrule
    \textbf{Full SurgQ-VLM (SFT+GRPO)} & --- & \textbf{0.779} & 0.799 & 0.818 & 0.820 & 0.645 & 0.741 & 0.803 \\
    \midrule
    \multicolumn{9}{@{}l}{\emph{Reward design}} \\
    \quad $-$ Answer-type dispatch ($R_1$ uniform)\,$\star$ & per-structure reward & 0.759 & 0.799 & 0.819 & 0.784 & 0.627 & 0.699 & 0.803 \\
    \midrule
    \multicolumn{9}{@{}l}{\emph{Architecture}} \\
    \quad $-$ Compressor (backbone-only) & dual-stream compressor & 0.746 & 0.785 & 0.808 & 0.742 & 0.611 & 0.656 & 0.707 \\
    \quad $-$ Question conditioning & GuideInjectors & 0.673 & 0.756 & 0.793 & 0.755 & 0.576 & 0.529 & 0.733 \\
    \midrule
    \multicolumn{9}{@{}l}{\emph{Training pipeline}} \\
    \quad $-$ GRPO (SFT only) & post-training stage & 0.754 & 0.799 & 0.818 & 0.769 & 0.632 & 0.690 & 0.803 \\
    \quad $-$ Q-type loss weighting & inverse-frequency & 0.744 & 0.767 & 0.763 & 0.812 & 0.628 & 0.702 & 0.787 \\
    \midrule
    \multicolumn{9}{@{}l}{\emph{Visual grounding}} \\
    \quad Text-only (blind, no visual input) & all frame tokens & 0.454 & 0.694 & 0.624 & 0.402 & 0.299 & 0.459 & 0.660 \\
    \bottomrule
  \end{tabular}
\end{table*}

Among fine-tuned baselines, SurgVidLM reaches 0.740 K-Acc (surgical specialist) and LLaVA-OneVision-7B 0.653 (strongest general video VLM); the gap reflects the compressor and answer-type-aware reward, not privileged training data. SurgVidLM samples 30 frames and 540 tokens; our model uses 45 frame instances yet produces 671 tokens and leads by 3.9\,pp K-Acc. The backbone-only row (same 45 frames, no compressor, \cref{tab:main}) leads SurgVidLM by only $0.6$\,pp, suggesting frame count is not the driver. The single-stream (slow-only) ablation conditions one stream with our GuideInjector --- HICom's scope~\cite{hicom} with mechanism, budget, and recipe held fixed, a controlled proxy rather than a reimplementation --- and reaches $0.750$ K-Acc vs.\ $0.754$ dual-stream at matched budget (finer temporal kernel: $648{+}23{=}671$ tokens). The headline margin is small (within the bootstrap CI half-width), but dual-streaming lifts Primary by $+0.9$\,pp with consistent per-group gains (binary $+1.4$, classification $+1.1$, counting $+1.5$\,pp) and enables the per-stream analysis of \cref{tab:stream}. A 3B backbone costs only $1.8$\,pp ($0.736$ vs.\ $0.754$), confirming the design is robust to backbone scale. Against the uncompressed backbone (7{,}605 post-merge tokens, $11{\times}$ our budget; $49{\times}$ counts the 32{,}805 pre-merge ViT patches), the compressor improves every metric.
The zero-shot medical RL model uAI-NEXUS MedVLM reaches 0.398 K-Acc on our test split --- a zero-shot, out-of-domain reference rather than a controlled comparison of reward designs; the uniform-$R_1$ ablation (\cref{sec:exp-ablation}) is the controlled evidence.

\subsection{Answer-structure gain analysis}
\label{sec:exp-perstructure}

\Cref{tab:perqtype} shows that the $+2.1$\,pp Primary gain does not distribute uniformly --- the asymmetry is the story. Binary and classification hold steady (changes ${<}0.1$\,pp) at 0.727 and 0.818; with four of five binary Q-types requiring transition or idle detection (chance 0.5), SFT cross-entropy is already a near-perfect proxy, leaving the reward no gap to close. The remaining three groups improve because SFT's token-level loss diverges from the evaluation metric.

\emph{Sentence-temporal} gains most ($+5.1$\,pp K-Acc): the reward's temporal term credits timing only within $\pm2$\,s --- a criterion invisible to cross-entropy. \emph{Multi-label} improves $+1.2$\,pp micro-F1 and $+5.1$\,pp keyword recall (K-Acc, \cref{tab:ablation}). \emph{Counting} adds $+2.6$\,pp on strict exact-match, with the reward's normalised count error improving in step (MAE $1.04\!\to\!0.93$: residual errors are predominantly off-by-one).

The one notable regression is Q13 (binary action-presence, $-3.9$\,pp), which pulls the binary group mean back to level. The reasoning-trace format GRPO rewards conflicts with the terse yes/no structure this Q-type expects, and the effect reproduces across GRPO configurations --- it is a genuine failure mode, not an artefact (mitigation: exempt binary Q-types from $R_2$; future work). A natural concern is verbosity: ROUGE-L is unchanged at $0.803$ (paired permutation $p{=}0.98$) while K-Acc, BLEU-4, and METEOR move at $p{<}0.0001$ --- the improvement is content, not length. Nor are gains confined to reward-aligned metrics: BLEU-4, METEOR, and CIDEr --- none of which enter the reward --- all improve (\cref{tab:main}). Within sentence-temporal, T-Acc is near-ceiling after SFT for all Q-types except Q08; the $+5.1$\,pp K-Acc gain is content-driven, concentrated in Q09 ($+8.2$\,pp; a point-lookup type whose T-Acc is vacuously saturated, supp.\ \S A.3) and Q08 ($+2.6$\,pp).

\subsection{Ablation studies}
\label{sec:exp-ablation}

\begin{figure*}[!t]
  \centering
  \includegraphics[width=\linewidth]{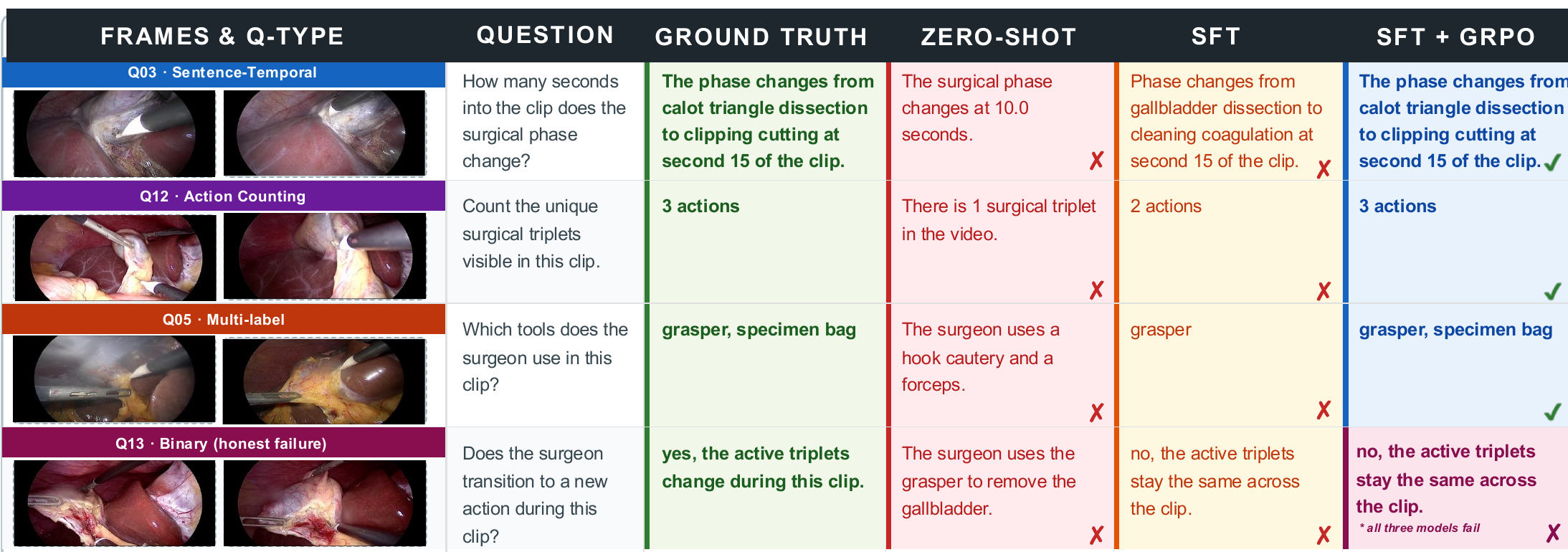}
  \caption{\textbf{Qualitative comparison} (sealed test set). Two frames, GT, ZS, SFT, and GRPO per row. Rows 1--3: GRPO recovers on Q03, Q12, and Q05 where ZS and SFT fail. Row 4: all models err on Q13 (failure case).}
  \label{fig:qual}
\end{figure*}

\Cref{tab:ablation} removes one component at a time; every removal lowers K-Acc below the full model's 0.779.

\vspace{-2pt}

\paragraph{Reward design.} Replacing the answer-type-aware reward with a single uniform $R_1$ ($\star$, the central contribution) --- the thresholded keyword-accuracy indicator applied to every Q-type --- drops K-Acc to 0.759, with the loss concentrated exactly on the structurally hard groups: multi-label $-3.6$\,pp and sentence-temporal $-4.2$\,pp, while binary and classification are unchanged. The uniform 0/1 reward collapses the continuous F1 and temporal signals: ${\sim}70\%$ of rollout groups then have zero reward variance and yield no gradient. The uniform reward recovers under a quarter of GRPO's gain --- the cleanest evidence that the reward \emph{design}, not GRPO itself, is what helps (full component ablations in the supplementary).

\vspace{-2pt}

\paragraph{Architecture.} Removing the compressor (backbone-only) lowers every group (K-Acc 0.746, multi-label 0.742, sentence-temporal 0.656). Removing question conditioning (GuideInjectors) yields a SlowFast-style question-agnostic dual-stream ablation~\cite{slowfastllava} at the same 671-token budget; sentence-temporal collapses to 0.529, the largest single-component drop, showing that per-question conditioning --- not dual-stream sampling alone --- preserves the relevant temporal evidence.

\begin{table}[h]
  \centering
  \footnotesize
  \setlength{\tabcolsep}{5pt}
  \caption{Stream specialisation: K-Acc change (pp) when each token group is zeroed at inference (7B+GRPO model). Most-critical component per row in \textbf{bold}. Most diagnostic Q-type shown per group; full breakdown in supplementary.}
  \label{tab:stream}
  \begin{tabular}{llccc}
    \toprule
    Question group & Q & $-$Slow & $-$Fast & $-$Global \\
    \midrule
    Phase          & Q03 & $+0.3$ & $\mathbf{-22}$ & $-0.7$ \\
    Instrument     & Q07 & $-0.5$ & $\mathbf{-2.6}$ & $+0.3$ \\
    Triplet/action & Q12 & $\mathbf{-2.1}$ & $-0.7$ & $-0.6$ \\
    Binary action  & Q13 & $-1.7$ & $-0.6$ & $\mathbf{-2.7}$ \\
    Temporal       & Q16 & $-2.3$ & $\mathbf{-12.7}$ & $+0.3$ \\
    \bottomrule
  \end{tabular}
\end{table}

\paragraph{Visual grounding.} The text-only blind baseline (\cref{tab:ablation}) scores $0.454$ K-Acc --- a $32.5$\,pp gap --- with the largest drops on multi-label ($-41.8$\,pp) and sentence-temporal ($-28.2$\,pp): the open-ended structures collapse without vision while label priors partially survive, so the hard-type scores reflect visual reasoning.

\paragraph{Training pipeline.} Dropping GRPO (SFT only) costs $2.5$\,pp K-Acc; GRPO training is stable across seeds (three runs: K-Acc $= 0.779 \pm 0.0003$, ${\sim}80{\times}$ below the gain; seeds share the SFT initialisation and one low-LR epoch; curves in supp.\ \S A.5). Disabling Q-type inverse-frequency loss weighting lowers binary and classification most ($-3.3$ and $-5.6$\,pp): the loss collapses onto high-count Q-types and under-trains the rare ones; the Q13 regression (\cref{sec:exp-perstructure}) recurs across all these variants.

\subsection{Stream specialisation and qualitative analysis}
\label{sec:exp-stream}
\label{sec:exp-qual}

Zeroing each token group at inference and re-measuring per Q-type (\cref{tab:stream}) reveals emergent specialisation: fast tokens dominate temporal Q-types (Q03 $-22.0$\,pp, Q16 $-12.7$\,pp), slow tokens dominate counting (Q12 $-2.1$ vs.\ $-0.7$\,pp fast), and global queries are load-bearing for holistic binary judgments (Q13 $-2.7$\,pp).

Qualitatively (\cref{fig:qual}), the zero-shot baseline emits fluent surgical English with hallucinated content, the SFT model produces well-formed answers with the wrong specifics, and GRPO recovers the correct content. Row~4 shows all three failing on the same hard binary example (the Q13 regression); Fig.~A2 in the supplementary adds six further examples.

\section{Limitations}  \textbf{(i) Scope:} all results are on one procedure (laparoscopic cholecystectomy); cross-procedure transfer is unmeasured. \textbf{(ii) Benchmark construction:} QA pairs come from deterministic templates over per-frame annotations, so linguistic diversity is bounded and scores may partly reflect template regularities; held-out-template and clinician-authored evaluation are future work. \textbf{(iii) Evaluation:} a few sentence-temporal question types saturate the $\pm2$\,s temporal-accuracy metric, so its group mean is optimistic. \textbf{(iv) Multi-step reasoning:} clip-level QA is atomic; cross-boundary questions are out of scope; multi-clip extension is future work. \textbf{(v) Annotation noise:} answers inherit label noise; disagreements concentrate near phase-transition boundaries~\cite{hong2021annotation, yamlahi2023selfdistill}.
\vspace{-3pt}
\section{Conclusion}
\label{sec:conclusion}

SurgQ-VLM shows that the bottleneck in surgical video QA is not decoder scale but the mismatch between a clip's heterogeneous questions and the uniform machinery that compresses and rewards them. Question-conditioned compression and answer-structure reward dispatch close that mismatch: the reward \emph{design} drives the gain --- over four times a uniform reward's --- concentrated on the open-ended structures where SFT plateaus.

\paragraph{Acknowledgments.}This work was funded by Centific Global Solutions, which also provided the compute resources.

{
    \small
    \bibliographystyle{ieeenat_fullname}
    \bibliography{refs}
}

\clearpage
\twocolumn[{\centering\Large\bfseries Supplementary Material\par\vspace{1.2em}}]
\setlength{\dblfloatsep}{6pt plus 2pt minus 2pt}
\renewcommand{\dbltopfraction}{0.95}
\renewcommand{\dblfloatpagefraction}{0.85}
\setcounter{dbltopnumber}{10}
\renewcommand{\topfraction}{0.95}
\renewcommand{\textfraction}{0.05}
\renewcommand{\floatpagefraction}{0.9}
\setcounter{topnumber}{4}
\setcounter{totalnumber}{6}
\noindent\textbf{Supplementary Content.}
In this supplementary material, we present the following:
\begin{itemize}[leftmargin=1.2em, itemsep=1pt, topsep=2pt]
  \item Dataset construction details: source annotations, clip segmentation, deterministic QA generation, answer-structure assignment, and a complete per-Q-type metadata table (\cref{app:dataset}).
  \item Question-type examples: one verbatim example per Q-type (Q01--Q18) in a single table, with ground-truth answers drawn directly from per-frame annotations (\cref{app:qtypes-examples}).
  \item Per-Q-type primary metric for SFT vs.\ GRPO across all 18 question types (\cref{app:perqtype-delta}).
  \item Full reward decomposition ablating each component of $R = R_1 + 0.3\,R_2 + 0.05\,R_3 - 0.3\,R_{\text{len}}$ (\cref{app:reward-full}).
  \item GRPO training stability: reward and KL-divergence curves over the full run, plus a group-size sensitivity check (\cref{app:reward-curves}).
  \item Stream specialisation: full per-Q-type K-Acc table under token-group zeroing, extending \cref{tab:stream} of the main paper (\cref{app:stream-full}).
  \item Temporal accuracy sensitivity at $\pm1$/$\pm2$/$\pm3$\,s tolerances on sentence-temporal Q-types (\cref{app:temporal}).
  \item MedGRPO-style per-group reward normalisation comparison (\cref{app:medgrpo-norm}).
  \item Additional qualitative examples (\cref{app:qual-extended}): a six-row figure with four rows (Q01, Q06, Q10, Q14) for binary and classification Q-types not shown in \cref{fig:qual}, plus two rows (Q16, Q08) for sentence-temporal Q-types where only GRPO succeeds; all examples from the sealed test set.
  \item Frame-count robustness: K-Acc under independent sweeps of the fast- and slow-stream frame budgets (\cref{app:framesweep}).
\end{itemize}

\appendix
\renewcommand{\thesection}{A.\arabic{section}}
\setcounter{table}{0}
\setcounter{figure}{0}
\renewcommand{\thetable}{A\arabic{table}}
\renewcommand{\thefigure}{A\arabic{figure}}
\renewcommand{\theHtable}{A\arabic{table}}
\renewcommand{\theHfigure}{A\arabic{figure}}
\section{Dataset Construction Details}
\label{app:dataset}

SurgQ-VLM is evaluated on a benchmark we built from two public laparoscopic cholecystectomy datasets. \textbf{Cholec80}~\cite{cholec80} provides per-frame phase and instrument annotations for 80 videos (7 phase labels; 7 instrument classes). \textbf{CholecT50}~\cite{cholect50} provides per-frame phase, instrument, verb, and triplet (\emph{instrument--verb--target}) annotations for 50 videos (same 7 phases; 10 verb classes; 6 instrument classes; 100 triplet combinations).

\paragraph{Clip segmentation.} Each video yields non-overlapping 30-second clips at 1\,FPS (30 frames per clip). \emph{Transition clips} are centred $\pm$15 frames around each detected phase or triplet transition; \emph{stable clips} are sampled every 45\,s inside single-phase regions that contain at least one active instrument. Clip centres are spaced at least 30\,s apart; videos shorter than 30\,s are skipped.

\paragraph{QA generation.} A deterministic rule-based generator reads per-frame annotation arrays (\texttt{phase\_per\_frame}, \texttt{instrument\_per\_frame}, \texttt{triplet\_per\_frame}, \texttt{verb\_per\_frame}) and fills question templates to produce fully grounded QA pairs. All answers are derived entirely from annotation values --- no language model is used for answer generation. Each Q-type has five surface-form variants (v0--v4) that ask the same underlying question with different surface phrasings; a variant is drawn uniformly at generation time to create lexically diverse training data while keeping ground-truth answers identical.

\paragraph{Answer-structure assignment.} Questions map to five evaluation structures:
\begin{itemize}[noitemsep,topsep=2pt]
  \item \textbf{Binary}: yes/no questions scored by normalised yes/no accuracy.
  \item \textbf{Classification}: single-label short-phrase scored by keyword accuracy (K-Acc).
  \item \textbf{Multi-label}: comma-separated set answers scored by micro-F1.
  \item \textbf{Counting}: integer-with-unit answers (``$N$ actions'') scored by exact-match.
  \item \textbf{Sentence-temporal}: free-text sentences embedding second values; scored by K-Acc (content) and T-Acc$_{\pm2\text{s}}$ (timing).
\end{itemize}

\paragraph{Video-level split.} All splits are at the video level to prevent clips from the same procedure appearing in both train and test. Because 45 CholecT50 videos originate from Cholec80, videos are canonicalised by their underlying surgery before partitioning: a CholecT50 video and its Cholec80 counterpart always share a split, so no surgery appears in more than one of train/validation/test (pairwise surgery-level overlap is zero). This yields Cholec80 48/9/23 and CholecT50 35/5/10 (train/val/test), with the CholecT50 test set matching the Rendezvous protocol~\cite{cholect50}. This yields train/validation/test sizes of 4{,}212 / 575 / 1{,}914 clips and 52{,}100 / 6{,}696 / 21{,}840 QA pairs respectively.

\paragraph{Annotation coverage.} Q01--Q09 and Q16--Q17 require only phase and instrument annotations and are generated for clips from both Cholec80 and CholecT50. Q10--Q15 and Q18 additionally require triplet or verb annotations, which are available only in CholecT50.

\Cref{tab:qtypes} lists all 18 question types with their group, answer structure, annotation source, temporal grounding requirement, and dataset coverage.

\begin{table*}[!t]
  \centering
  \footnotesize
  \setlength{\tabcolsep}{3pt}
  \caption{All 18 question types. \emph{Coverage}: Both = Cholec80 + CholecT50; T50 = CholecT50 only. \emph{Temporal}: question or answer references a specific second value.}
  \label{tab:qtypes}
  \begin{tabular}{lllllcc}
    \toprule
    Q & Name & Group & Answer struct. & Annotation source & Temporal & Coverage \\
    \midrule
    Q01 & Current phase              & Phase      & Classification & phase majority vote               & No  & Both \\
    Q02 & Phase change               & Phase      & Binary         & phase change flag                 & No  & Both \\
    Q03 & Phase transition time      & Phase      & Sent.-temporal & phase segment boundary second     & Yes & Both \\
    Q04 & Phase sequence             & Phase      & Sent.-temporal & phase segments, ordered           & Yes & Both \\
    Q05 & Active instrument list     & Instrument & Multi-label    & instrument union across clip      & No  & Both \\
    Q06 & Instrument presence        & Instrument & Binary         & per-instrument membership         & No  & Both \\
    Q07 & Instrument handoff         & Instrument & Binary         & instrument transition events      & Yes & Both \\
    Q08 & Instrument active duration & Instrument & Sent.-temporal & per-instrument frame count        & Yes & Both \\
    Q09 & Instrument at second $N$   & Instrument & Sent.-temporal & point lookup in frame array       & Yes & Both \\
    Q10 & Primary triplet            & Triplet    & Classification & triplet frame-count majority      & No  & T50  \\
    Q11 & Action sequence            & Triplet    & Multi-label    & triplet segments, ordered         & No  & T50  \\
    Q12 & Distinct action count      & Triplet    & Counting       & distinct triplet elements         & No  & T50  \\
    Q13 & Action presence            & Triplet    & Binary         & triplet change flag               & No  & T50  \\
    Q14 & Anatomical target          & Triplet    & Classification & triplet target component          & No  & T50  \\
    Q15 & Primary verb               & Triplet    & Classification & verb majority vote                & No  & T50  \\
    Q16 & Instrument at phase change & Temporal   & Sent.-temporal & phase boundary $+$ instrument arr.\ & Yes & Both \\
    Q17 & Instrument idle period     & Temporal   & Binary         & instrument idle segments          & Yes & Both \\
    Q18 & Action before/after change & Temporal   & Sent.-temporal & triplet segments before/after     & Yes & T50  \\
    \bottomrule
  \end{tabular}
\end{table*}

\section{Question-Type Examples}
\label{app:qtypes-examples}

\Cref{tab:qtype-examples} shows one representative example per question type, drawn verbatim from the sealed test split. Ground-truth answers are derived directly from per-frame annotations with no language model involvement.

\begin{table*}[!t]
  \centering
  \footnotesize
  \setlength{\tabcolsep}{4pt}
  \caption{One representative example per Q-type (Q01--Q18), verbatim from the sealed test split. Q03/Q04 are from the same clip. Q10--Q15 and Q18 require CholecT50 triplet annotations. Q07 gold answers embed a second value but the Q-type is scored by binary accuracy; Q08 is scored by T-Acc$_{\pm2\text{s}}$. Q09 answers are instrument lists with no second embedded, so their T-Acc is vacuously 1 (\cref{app:perqtype-delta}).}
  \label{tab:qtype-examples}
  \begin{tabular}{p{0.07\linewidth}lp{0.34\linewidth}p{0.40\linewidth}}
    \toprule
    Group & Q & Example question & Ground-truth answer \\
    \midrule
    \multirow{4}{=}{\textit{Phase}}
    & Q01 & Which phase of the cholecystectomy procedure is shown here?
           & The clip primarily shows calot triangle dissection. \\[3pt]
    & Q02 & Does the surgical phase change at any point during this clip?
           & No, the surgical phase remains constant throughout this clip. \\[3pt]
    & Q03 & How many seconds into the clip does the surgical phase change?
           & The phase changes from gallbladder packaging to cleaning coagulation at second 18 of the clip. \\[3pt]
    & Q04 & Which phases appear in this clip and in what order?
           & gallbladder packaging from second 0 to second 17, then cleaning coagulation from second 18 to the end of the clip. \\
    \midrule
    \multirow{5}{=}{\textit{Instrument}}
    & Q05 & What instruments are active at any point in this clip?
           & hook, grasper \\[3pt]
    & Q06 & Is the grasper used at any point in this clip?
           & Yes, the grasper is used in this clip. \\[3pt]
    & Q07 & Does the active surgical tool change during this video?
           & Yes, the active instrument changes from hook to grasper at second 15. \\[3pt]
    & Q08 & How long is the hook active during this clip?
           & The hook is used for 29 seconds, from second 0 to second 29 of the clip. \\[3pt]
    & Q09 & What is the active instrument 13 seconds into this clip?
           & hook \\
    \midrule
    \multirow{6}{=}{\textit{Triplet}}
    & Q10 & Which action dominates this clip?
           & The primary action in this clip is grasper grasp gallbladder. \\[3pt]
    & Q11 & Walk me through the sequence of actions in this clip.
           & The actions in this clip are, in order of first appearance: grasper retract gallbladder, hook dissect cystic duct. \\[3pt]
    & Q12 & How many distinct surgical actions occur in this clip?
           & 2 actions \\[3pt]
    & Q13 & Is there an action transition visible in this clip?
           & Yes, the active triplets change during this clip. \\[3pt]
    & Q14 & Identify the main target of the surgical actions here.
           & The primary anatomical target in this clip is gallbladder. \\[3pt]
    & Q15 & What is the primary surgical maneuver in this clip?
           & The primary surgical maneuver in this clip is grasp. \\
    \midrule
    \multirow{3}{=}{\textit{Temporal}}
    & Q16 & How does the surgical instrument change when the phase transitions in this clip?
           & When the phase changes from calot triangle dissection to clipping cutting at second 3, the grasper is replaced by the clipper. \\[3pt]
    & Q17 & Are there seconds in this clip where no instrument is being used?
           & Yes, there is a period of 29 seconds with no active instrument, from second 0 to second 28 of the clip. \\[3pt]
    & Q18 & What surgical action appears immediately after the main transition in this clip?
           & Before the transition the action is hook dissect cystic plate; after the transition the action is grasper retract gallbladder. \\
    \bottomrule
  \end{tabular}
\end{table*}

\section{Per-Q-Type Primary Metric: SFT vs.\ GRPO}
\label{app:perqtype-delta}

\Cref{tab:perqtype-delta} is the full per-Q-type breakdown behind the group aggregate in \cref{tab:perqtype} of the main paper, each Q-type scored on its group's primary metric (binary accuracy for binary types; K-Acc for classification; micro-F1 for multi-label; exact-match for counting; T-Acc$_{\pm2\text{s}}$ for sentence-temporal). Gains concentrate in the open-ended types --- multi-label (Q05), counting (Q12), and sentence-temporal generation (Q08); binary and classification Q-types are at or near SFT ceiling. The one notable regression is Q13 (binary action-presence, $-3.85$\,pp); the main paper analyses this failure mode in detail. Q09 and Q18 score 1.0000 on T-Acc because their gold answers contain no explicit second values, so the temporal check is vacuously satisfied; treat those rows as a metric caveat, not as evidence of perfect temporal grounding.

\begin{table}[!htbp]
  \centering
  \footnotesize
  \setlength{\tabcolsep}{4pt}
  \caption{Per-Q-type primary metric, SFT vs.\ GRPO (sealed test split, $n{=}21{,}840$). Metric per group: binary accuracy (bin), classification K-Acc (cls), multi-label F1 (mul), counting exact-match (cnt), sentence-temporal T-Acc$_{\pm2\text{s}}$ (s-tmp).}
  \label{tab:perqtype-delta}
  \begin{tabular}{llccr}
    \toprule
    Q & Name (group) & SFT & GRPO & $\Delta$\,(pp) \\
    \midrule
    Q01 & Current phase (cls)               & 0.8163 & 0.8186 & $+0.23$ \\
    Q02 & Phase change (bin)                & 0.6694 & 0.6612 & $-0.82$ \\
    Q03 & Phase transition time (s-tmp)     & 0.8793 & 0.8793 & $0.00$ \\
    Q04 & Phase sequence (s-tmp)            & 0.9253 & 0.9253 & $0.00$ \\
    Q05 & Active instrument list (mul)      & 0.7738 & 0.7896 & $+1.58$ \\
    Q06 & Instrument presence (bin)         & 0.8623 & 0.8623 & $0.00$ \\
    Q07 & Instrument handoff (bin)          & 0.5145 & 0.5182 & $+0.37$ \\
    Q08 & Instrument active duration (s-tmp) & 0.7271 & 0.7606 & $+3.35$ \\
    Q09 & Instrument at second $N$ (s-tmp)  & 1.0000 & 1.0000 & $0.00$ \\
    Q10 & Primary triplet (cls)             & 0.7284 & 0.7190 & $-0.94$ \\
    Q11 & Action sequence (mul)             & 0.7385 & 0.7363 & $-0.22$ \\
    Q12 & Distinct action count (cnt)       & 0.3284 & 0.3542 & $+2.58$ \\
    Q13 & Action presence (bin)             & 0.6250 & 0.5865 & $-3.85$ \\
    Q14 & Anatomical target (cls)           & 0.8765 & 0.8762 & $-0.03$ \\
    Q15 & Primary verb (cls)                & 0.8570 & 0.8545 & $-0.25$ \\
    Q16 & Instrument at phase change (s-tmp) & 0.9028 & 0.9028 & $0.00$ \\
    Q17 & Instrument idle period (bin)      & 0.5915 & 0.5922 & $+0.07$ \\
    Q18 & Action before/after change (s-tmp) & 1.0000 & 1.0000 & $0.00$ \\
    \bottomrule
  \end{tabular}
\end{table}

\section{Full Reward Decomposition}
\label{app:reward-full}

\Cref{tab:reward-full} extends the $R_1$-uniform ablation in \cref{tab:ablation} of the main paper by independently ablating each reward component. The anchor is the full reward $R = R_1 + 0.3\,R_2 + 0.05\,R_3 - 0.3\,R_{\text{len}}$; each row strips or substitutes one term and retrains from the SFT checkpoint with the same one-epoch GRPO~\cite{grpo} budget (DAPO-style asymmetric clipping~\cite{dapo}) on the sealed test split ($n{=}21{,}840$).

Stripping any single component causes a ${\approx}1.9$\,pp drop in K-Acc. The effect is concentrated where it matters: multi-label K-Acc drops from 0.820 (full) to 0.786--0.788, and sentence-temporal K-Acc from 0.741 to 0.700--0.703 --- these are the two answer structures where GRPO provides the most lift over SFT. Binary and classification Q-types are at SFT ceiling and unaffected. Q-type inverse-frequency weighting provides the smallest headline gain ($1.4$\,pp on K-Acc) but is complementary: without it, multi-label and sentence-temporal improvements are also reduced.

\begin{table*}[!t]
  \centering
  \small
  \caption{Full reward decomposition on the sealed test split ($n{=}21{,}840$). Each row removes or replaces one component from the full GRPO reward; all other hyperparameters unchanged. Group columns report per-group K-Acc.}
  \label{tab:reward-full}
  \begin{tabular}{lcccccc}
    \toprule
    Variant & K-Acc & Binary & Classif. & Multi-lbl & Counting & S-Temp \\
    \midrule
    \textbf{Full} $R = R_1 + 0.3\,R_2 + 0.05\,R_3 - 0.3\,R_{\text{len}}$ & \textbf{0.779} & \textbf{0.799} & \textbf{0.818} & \textbf{0.820} & \textbf{0.645} & \textbf{0.741} \\
    \midrule
    $R_1$ only (strip $R_2$, $R_3$, $R_{\text{len}}$)   & 0.760 & 0.799 & 0.820 & 0.788 & 0.630 & 0.700 \\
    $R_1 + R_2$ (strip $R_3$, $R_{\text{len}}$)          & 0.761 & 0.799 & 0.820 & 0.786 & 0.633 & 0.703 \\
    $R_1 + R_3$ (strip $R_2$, $R_{\text{len}}$)          & 0.760 & 0.799 & 0.819 & 0.788 & 0.629 & 0.701 \\
    No length penalty ($R_{\text{len}} = 0$)              & 0.760 & 0.798 & 0.820 & 0.787 & 0.636 & 0.701 \\
    No Q-type IFW in reward (uniform weighting)           & 0.765 & 0.799 & 0.820 & 0.794 & 0.628 & 0.712 \\
    \bottomrule
  \end{tabular}
\end{table*}

\section{GRPO Training Stability}
\label{app:reward-curves}

\begin{figure}[!h]
  \centering
  \includegraphics[width=\linewidth]{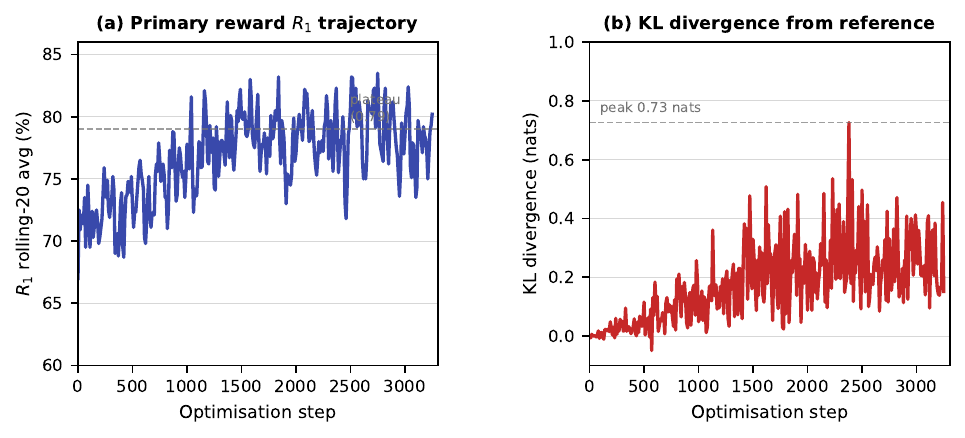}
  \caption{GRPO training dynamics (seed 42, 3{,}256 steps). \textbf{Left:} primary reward $R_1$ (20-step rolling average) rises then plateaus at ${\approx}0.79$. \textbf{Right:} KL divergence from the reference policy stays bounded (peak 0.73\,nats) with no collapse. Both curves confirm stable optimisation without reward hacking.}
  \label{fig:reward-curves}
\end{figure}

\Cref{fig:reward-curves} shows the primary reward $R_1$ and KL divergence over the full 3{,}256 optimisation steps of the reported GRPO run (seed\,42; 20-step rolling averages). $R_1$ rises from 0.70 to a plateau of ${\approx}0.79$ by step 1{,}600 and remains stable through training --- no monotone increase that would indicate reward hacking. KL divergence stays bounded throughout, with a peak of 0.73\,nats (step 2{,}381) and a mean of 0.27\,nats over the final 1{,}256 steps; there is no collapse or divergence. GRPO hyperparameters: group size $G{=}8$, KL coefficient $\beta{=}0.04$, DAPO asymmetric clipping $\epsilon_{\mathrm{low}}{=}0.2$, $\epsilon_{\mathrm{high}}{=}0.28$. A group-size sensitivity run ($G{=}4$, otherwise identical configuration) reaches K-Acc~$0.775$ vs.\ $0.779$ for $G{=}8$ ($-0.4$\,pp on the sealed test split of 21{,}840 pairs), confirming that the GRPO gain is robust to group size within this range.

\section{Stream Ablation: Per-Q-Type Breakdown}
\label{app:stream-full}

\Cref{tab:stream-full} is the full per-Q-type expansion of \cref{tab:stream} in the main paper, reporting K-Acc under each inference-time stream-zeroing condition for the 7B\,+\,GRPO model. The largest drop per row is highlighted by sign and magnitude. Q06 achieves K-Acc\,=\,1.000 under every condition --- this reflects that the model reliably emits a yes/no keyword regardless of which stream is zeroed (K-Acc detects keyword presence), not that all answers are polarity-correct; binary accuracy on Q06 is 0.862 per \cref{tab:perqtype-delta}.

\begin{table*}[!t]
  \centering
  \small
  \caption{Stream zeroing, per-Q-type K-Acc (7B\,+\,GRPO, sealed test split). Each $\Delta$ column is that condition minus Full. Largest-magnitude $\Delta$ per row in \textbf{bold}.}
  \label{tab:stream-full}
  \begin{tabular}{lcccc|rrr}
    \toprule
    Q & Full & $-$slow & $-$fast & $-$global & $\Delta$\,slow & $\Delta$\,fast & $\Delta$\,global \\
    \midrule
    Q01 & 0.8201 & 0.8199 & 0.7932 & 0.8194 & $-0.0002$ & $-0.0269$ & $-0.0007$ \\
    Q02 & 0.6694 & 0.6557 & 0.5082 & 0.6585 & $-0.0137$ & $\mathbf{-0.1612}$ & $-0.0109$ \\
    Q03 & 0.7680 & 0.7706 & 0.5485 & 0.7611 & $+0.0026$ & $\mathbf{-0.2195}$ & $-0.0069$ \\
    Q04 & 0.8285 & 0.8293 & 0.7350 & 0.8289 & $+0.0008$ & $\mathbf{-0.0935}$ & $+0.0004$ \\
    Q05 & 0.8276 & 0.8236 & 0.8201 & 0.8275 & $-0.0040$ & $-0.0075$ & $-0.0001$ \\
    Q06 & 1.0000 & 1.0000 & 1.0000 & 1.0000 & $0.0000$ & $0.0000$ & $0.0000$ \\
    Q07 & 0.4567 & 0.4515 & 0.4307 & 0.4593 & $-0.0052$ & $-0.0260$ & $+0.0026$ \\
    Q08 & 0.7096 & 0.7129 & 0.7098 & 0.7031 & $+0.0033$ & $+0.0002$ & $-0.0065$ \\
    Q09 & 0.7591 & 0.7522 & 0.7512 & 0.7589 & $-0.0069$ & $-0.0079$ & $-0.0002$ \\
    Q10 & 0.7199 & 0.7149 & 0.6999 & 0.7181 & $-0.0050$ & $-0.0200$ & $-0.0018$ \\
    Q11 & 0.7915 & 0.7900 & 0.7841 & 0.7914 & $-0.0015$ & $-0.0074$ & $-0.0001$ \\
    Q12 & 0.6485 & 0.6273 & 0.6411 & 0.6430 & $\mathbf{-0.0212}$ & $-0.0074$ & $-0.0055$ \\
    Q13 & 0.8385 & 0.8212 & 0.8327 & 0.8115 & $-0.0173$ & $-0.0058$ & $\mathbf{-0.0270}$ \\
    Q14 & 0.8762 & 0.8767 & 0.8651 & 0.8765 & $+0.0005$ & $-0.0111$ & $+0.0003$ \\
    Q15 & 0.8532 & 0.8433 & 0.8265 & 0.8520 & $-0.0099$ & $-0.0267$ & $-0.0012$ \\
    Q16 & 0.8002 & 0.7770 & 0.6731 & 0.8033 & $\mathbf{-0.0232}$ & $\mathbf{-0.1271}$ & $+0.0031$ \\
    Q17 & 0.6204 & 0.6138 & 0.5927 & 0.6211 & $-0.0066$ & $-0.0277$ & $+0.0007$ \\
    Q18 & 0.8097 & 0.8120 & 0.8023 & 0.8097 & $+0.0023$ & $-0.0074$ & $0.0000$ \\
    \bottomrule
  \end{tabular}
\end{table*}

\section{Temporal Accuracy Sensitivity}
\label{app:temporal}

\Cref{tab:temporal-sensitivity} reports Temporal Accuracy at $\pm 1$\,s, $\pm 2$\,s, and $\pm 3$\,s tolerances on the six sentence-temporal Q-types (Q03, Q04, Q08, Q09, Q16, Q18). The monotonic rise across tolerances rules out random guessing: errors are tightly clustered near ground truth. Q09 and Q18 score 1.0000 on this metric for both SFT and GRPO because their gold answers contain no explicit second values, so the temporal check is vacuously satisfied; those rows should be read as a caveat on the metric, not as evidence of perfect temporal grounding (see also \cref{app:perqtype-delta}).

\smallskip\noindent\textit{Note on values vs.\ main paper.}~~The T-Acc$_{\pm2\text{s}}$ values here (SFT\,0.874, GRPO\,0.888) differ from the sentence-temporal row in \cref{tab:perqtype} of the main paper (0.690\,/\,0.741) because they are \emph{different metrics}. The main paper's aggregate dispatches sentence-temporal types to keyword accuracy (K-Acc) as the group primary --- the fraction of predictions containing a correct keyword or phrase --- and \cref{tab:perqtype} is labelled accordingly. This table reports T-Acc$_{\pm2\text{s}}$ (second-extraction matching), which is higher because the sentence templates reliably produce second values; K-Acc is the conservative headline since it reflects overall response correctness.

\begin{table}[!htbp]
  \centering
  \small
  \caption{Temporal accuracy at $\pm 1$/$\pm 2$/$\pm 3$\,s on sentence-temporal Q-types ($n{=}8{,}210$ sentence-temporal test QA pairs).}
  \label{tab:temporal-sensitivity}
  \begin{tabular}{lccc}
    \toprule
    Model        & $\pm 1$\,s & $\pm 2$\,s & $\pm 3$\,s \\
    \midrule
    7B SFT       & 0.860 & 0.874 & 0.884 \\
    \textbf{7B\,+\,GRPO} & \textbf{0.876} & \textbf{0.888} & \textbf{0.899} \\
    $\Delta$     & $+0.016$ & $+0.014$ & $+0.015$ \\
    \bottomrule
  \end{tabular}
\end{table}

\section{MedGRPO-Style Per-Group Reward Normalisation}
\label{app:medgrpo-norm}

MedGRPO~\cite{medgrpo} normalises rewards across heterogeneous medical datasets with a median/IQR sigmoid, $r \mapsto \sigma\!\big(3(r-\mathrm{med}_g)/\mathrm{IQR}_g\big)$, so imbalanced reward scales do not collapse multi-dataset training. We adapt this to operate \emph{per answer-structure group} within our single procedure, with a saturated-group skip when $\mathrm{IQR}_g<0.1$ (which keeps already-converged groups from mapping every rollout to $0.5$). Warm-started from the SFT checkpoint with the same one-epoch budget as our GRPO run, the normalised variant reaches the same headline K-Acc as answer-type dispatch alone (\cref{tab:medgrpo-norm}); per-group differences are within $\pm 1$\,pp in either direction. The parity confirms the paper's argument: the cross-dataset reward heterogeneity that normalisation targets is absent within a single procedure, where per-group reward scales are already comparable.

\begin{table}[!htbp]
  \centering
  \small
  \caption{Answer-type dispatch (our full model) vs.\ MedGRPO-style per-group reward normalisation~\cite{medgrpo}, both warm-started from the SFT checkpoint with a matched one-epoch budget (sealed test split, $n{=}21{,}840$).}
  \label{tab:medgrpo-norm}
  \begin{tabular}{lcc}
    \toprule
    Metric & Dispatch (ours) & $+$ per-group norm. \\
    \midrule
    K-Acc            & 0.779 & 0.779 \\
    Primary          & 0.743 & 0.743 \\
    ROUGE-L          & 0.803 & 0.811 \\
    \midrule
    \multicolumn{3}{@{}l}{\emph{Per-group K-Acc $\Delta$ (norm.\ $-$ dispatch)}} \\
    Binary           & --- & $-0.3$ \\
    Classification   & --- & $-0.2$ \\
    Multi-label      & --- & $+0.3$ \\
    Counting         & --- & $-0.9$ \\
    Sentence-temporal & --- & $+0.3$ \\
    \bottomrule
  \end{tabular}
\end{table}

\section{Additional Qualitative Examples}
\label{app:qual-extended}

\begin{figure*}[!t]
  \centering
  \includegraphics[width=\linewidth]{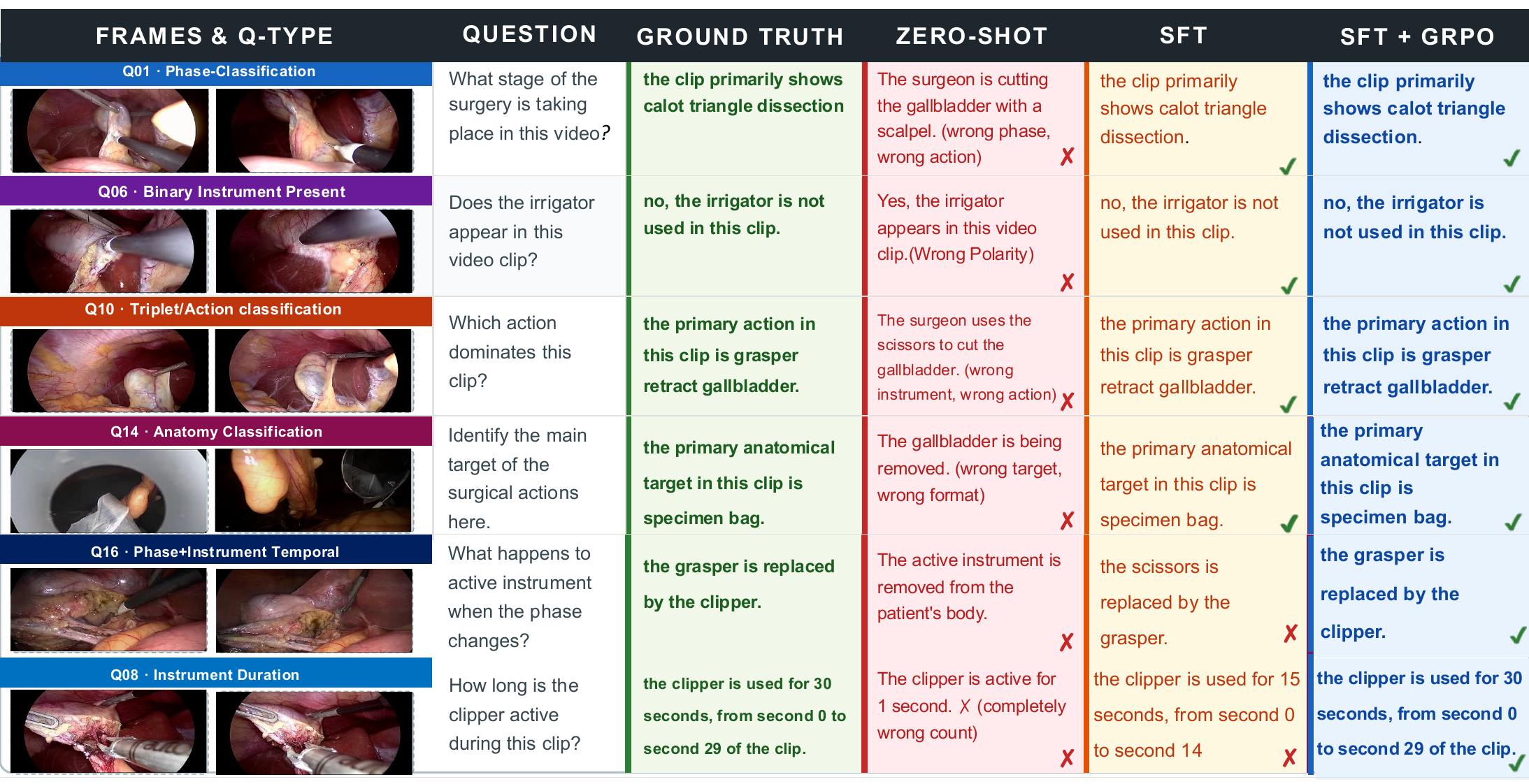}
  \caption{Additional qualitative examples across six Q-types (sealed test set). Rows 1--4 cover binary and classification Q-types (Q01, Q06, Q10, Q14); rows 5--6 cover sentence-temporal Q-types (Q16, Q08). Each row shows two representative frames, the question, ground truth, and predictions from zero-shot (ZS), SFT, and SFT$+$GRPO. ZS fails on all six through systematic errors --- wrong phase names, wrong polarity, wrong instrument or action, vague temporal descriptions with no named content. For the binary and classification rows, SFT already matches the ground truth and GRPO adds no further gain, consistent with the near-zero deltas in \cref{tab:perqtype} of the main paper. For the sentence-temporal rows, SFT learns the surface format but fills it with wrong phase names, wrong instruments, or wrong durations; GRPO recovers the fully correct structured answer in both cases.}
  \label{fig:qual-extended}
\end{figure*}

\Cref{fig:qual-extended} complements \cref{fig:qual} of the main paper, which focused on generation-heavy Q-types (phase-transition timing Q03, action counting Q12, multi-label instrument lists Q05). Rows 1--4 cover binary and classification Q-types not shown in the main figure: four sealed test-set clips where zero-shot fails through systematic errors (wrong phase, wrong polarity, wrong action, wrong anatomy), SFT reaches the correct answer, and GRPO adds no further gain. Rows 5--6 extend coverage to sentence-temporal Q-types (Q16: phase-and-instrument-at-transition; Q08: instrument active duration), where the answer requires precise phase names, instrument identities, and an exact second value: zero-shot produces vague or structurally wrong output, SFT learns the sentence template but fills it with wrong content, and GRPO recovers the fully correct structured answer.

\section{Frame-Count Robustness}
\label{app:framesweep}

\begin{figure*}[!t]
  \centering
  \includegraphics[width=\linewidth]{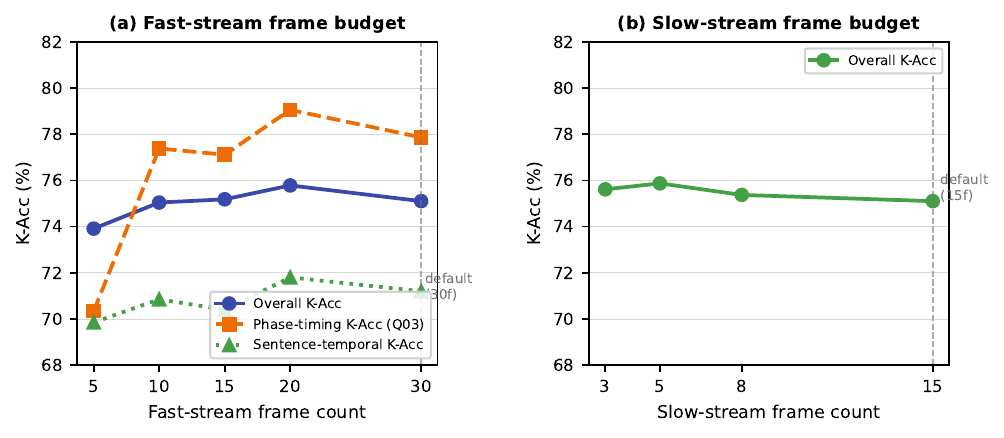}
  \caption{Frame-count robustness. \emph{(a)} Fast stream (slow=15 fixed): Q03 drops 7.5\,pp at fast=5; K-Acc flat above 10 frames. \emph{(b)} Slow stream (fast=30 fixed): K-Acc varies by only 0.77\,pp across 3--15 frames.}
  \label{fig:framesweep}
\end{figure*}

\Cref{fig:framesweep} reports frame-count robustness, varying the fast and slow stream frame budgets independently while holding the other at its default. The fast stream is sensitive exactly where the architecture assigns it: phase-transition timing (Q03) rises $7.5$\,pp as fast frames increase from 5 to 10, while overall K-Acc plateaus above 10 frames. By contrast, varying the slow-stream budget between 3 and 15 frames moves K-Acc by only $0.77$\,pp: slow-stream tokens carry redundant spatial context at this density, and the fast stream is the load-bearing path for temporal Q-types.

\end{document}